\def\paperID{0000}
\def\confName{ICCV}
\def\confYear{2025}

\def\paperTitle{
\textbf{\textcolor{blue}{VLA-LPAF}: \textcolor{blue}{L}ightweight \textcolor{blue}{P}erspective-\textcolor{blue}{A}daptive \textcolor{blue}{F}usion for \textcolor{blue}{V}ision-\textcolor{blue}{L}anguage-\textcolor{blue}{A}ction to Enable More Unconstrained Robotic Manipulation}
}

\def\authorBlock{
    Jinyue Bian \qquad
    Zhaoxing Zhang \qquad
    Zhengyu Liang \qquad
    Shiwei Zheng \\
    Shengtao Zhang \qquad
    Rong Shen \qquad
    Chen Yang \qquad
    Anzhou Hou \\
    China, Beijing, Li Auto Inc. \\
    {\tt\small \{bianjinyue, zhangzhaoxing, liangzhengyu, zhengshiwei\}@lixiang.com}\\
    {\tt\small \{zhangshengtao, shenrong, yangchen11, houanzhou\}@lixiang.com}
}

\newif\ifreview 
\newif\ifarxiv \newcommand{\arxiv}{\arxivtrue}
\newif\ifcamera 
\newif\ifrebuttal 

\arxiv

\pdfoutput=1
\documentclass[10pt,twocolumn,letterpaper]{article}
\ifreview \usepackage[review]{cvpr} \fi
\ifarxiv \usepackage[pagenumbers]{cvpr} \fi
\ifrebuttal \usepackage[rebuttal]{cvpr} \fi
\ifcamera \usepackage{cvpr} \fi

\usepackage{graphicx}	
\usepackage{amsmath}	
\usepackage{amssymb}	
\usepackage{booktabs}
\usepackage{times}
\usepackage{microtype}
\usepackage{epsfig}
\usepackage{caption}
\usepackage{float}
\usepackage{placeins}
\usepackage{color, colortbl}
\usepackage{stfloats}
\usepackage{enumitem}
\usepackage{tabularx}
\usepackage{xstring}
\usepackage{multirow}
\usepackage{xspace}
\usepackage{url}
\usepackage{subcaption}
\usepackage{xcolor}
\usepackage[hang,flushmargin]{footmisc}

\ifcamera \usepackage[accsupp]{axessibility} \fi

% ...

\ifarxiv  \fi

\newcommand{\R}[1]{{%
    \textbf{%
        \ifstrequal{#1}{1}{\textcolor{red}{R#1}}{%
        \ifstrequal{#1}{2}{\textcolor{blue}{R#1}}{%
        \ifstrequal{#1}{3}{\textcolor{magenta}{R#1}}{%
        \ifstrequal{#1}{4}{\textcolor{teal}{R#1}}{%
                           \textcolor{cyan}{R#1}%
        }}}}%
    }%
}}

\usepackage{xr-hyper}

\makeatletter
\newcommand*{\addFileDependency}[1]{
  \typeout{(#1)}
  \@addtofilelist{#1}
  \IfFileExists{#1}{}{\typeout{No file #1.}}
}

\makeatother
\newcommand*{\myexternaldocument}[1]{
    \externaldocument{#1}
    \addFileDependency{#1.tex}
    \addFileDependency{#1.aux}
}

\definecolor{cvprblue}{rgb}{0.21,0.49,0.74}
\usepackage[pagebackref,breaklinks,colorlinks,allcolors=cvprblue]{hyperref}
\usepackage[capitalize]{cleveref}
\crefname{section}{Sec.}{Secs.}
\crefname{table}{Table}{Tables}
\crefname{figure}{Fig.}{Figs.}

\ifarxiv \crefname{appendix}{App.}{Apps.}
\else \crefname{appendix}{Suppl.}{Suppls.} \fi

\unless\ifarxiv \myexternaldocument{_supplementary} \fi

\begin{document}
\title{\paperTitle}
\author{\authorBlock}
\maketitle
\begin{abstract}
The Visual-Language-Action (VLA) models can follow text instructions according to visual observations of the surrounding environment. This ability to map multimodal inputs to actions is derived from the training of the VLA model on extensive standard demonstrations. These visual observations captured by third-personal global and in-wrist local cameras are inevitably varied in number and perspective across different environments, resulting in significant differences in the visual features. This perspective heterogeneity constrains the generality of VLA models. In light of this, we first propose the lightweight module \textbf{VLA-LPAF} to foster the perspective adaptivity of VLA models using only 2D data. \textbf{VLA-LPAF} is finetuned using images from a single view and fuses other multiview observations in the latent space, which effectively and efficiently bridge the gap caused by perspective inconsistency. We instantiate our VLA-LPAF framework with the VLA model RoboFlamingo to construct \textbf{RoboFlamingo-LPAF}. Experiments show that \textbf{RoboFlamingo-LPAF} averagely achieves around \textbf{8\%} task success rate improvement on CALVIN, \textbf{15\%} on LIBERO, and \textbf{30\%} on a custom simulation benchmark. We also demonstrate the view-adaptive characteristics developed of the proposed \textbf{RoboFlamingo-LPAF} through real-world tasks.
\end{abstract}
\section{Introduction}

\begin{figure}[!ht]
\centering
\captionsetup{justification=justified, singlelinecheck=false}
\includegraphics[width=0.95\columnwidth]{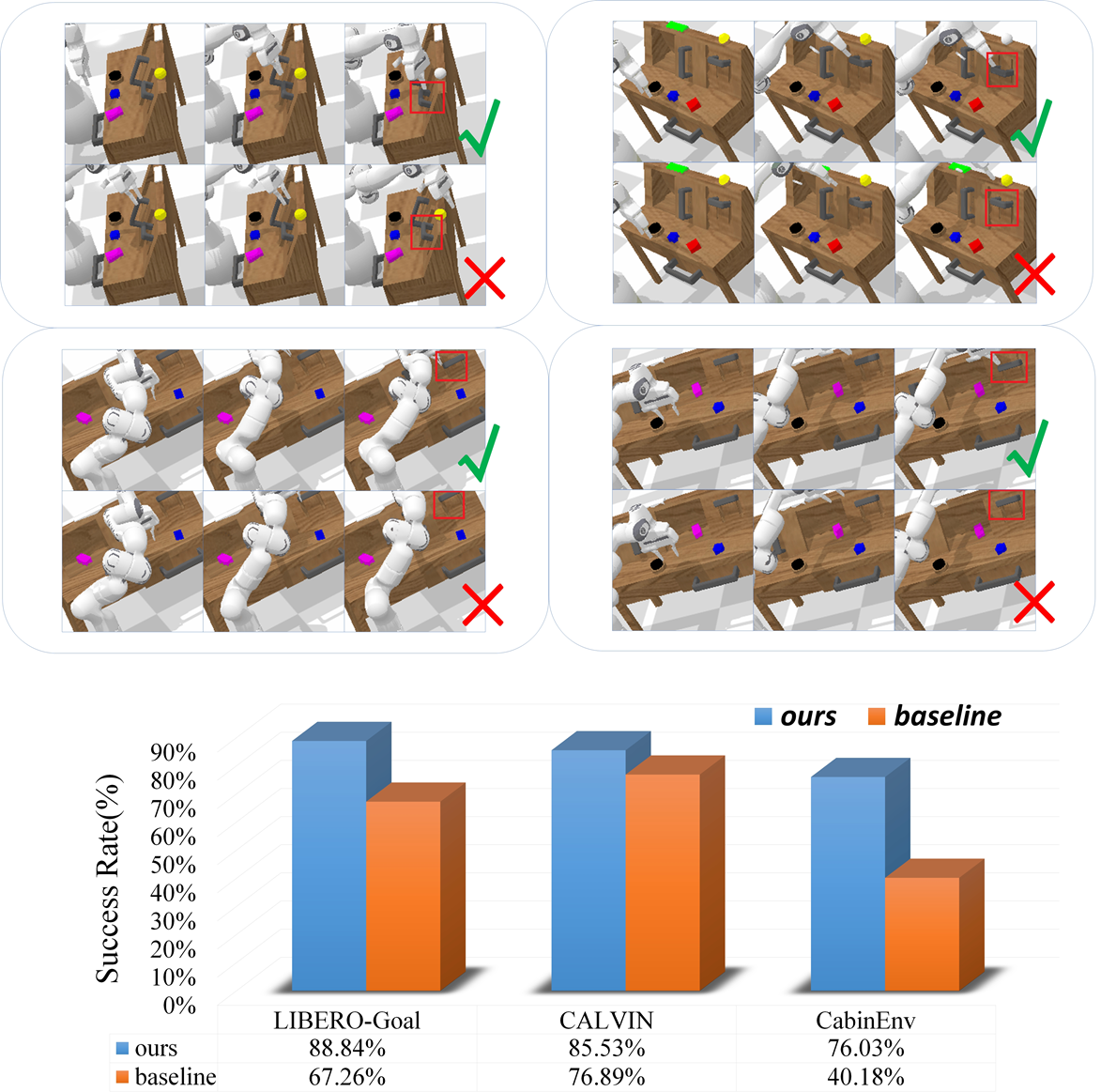}
\caption{We generate various 2D perspectives in CALVIN simulator and the comparison results demonstrate that the VLA model RoboFlamingo is capable to perform the tasks if constructed under our VLA-LPAF framework. We mark the task execution result by red rectangles for the four examples in the top part and statistics across three datasets in the bottom to show the advantage of our method.}
\label{lv-i0}
\end{figure}

Visual-Language-Action (VLA) models \cite{li2024visionlanguagefoundationmodelseffective,kim2024openvlaopensourcevisionlanguageactionmodel,kim2025finetuningvisionlanguageactionmodelsoptimizing,yan2024robommallinonemultimodallarge,wu2023unleashinglargescalevideogenerative,cheang2024gr2generativevideolanguageactionmodel,cheang2025gr3technicalreport,black2024pi0visionlanguageactionflowmodel,intelligence2025pi05visionlanguageactionmodelopenworld,goyal2024rvt2learningprecisemanipulation,goyal2023rvtroboticviewtransformer,sun2025geovlaempowering3drepresentations,li2025bridgevlainputoutputalignmentefficient,wen2025diffusionvlageneralizableinterpretablerobot,qu2025spatialvlaexploringspatialrepresentations,brohan2023rt1roboticstransformerrealworld,brohan2023rt2visionlanguageactionmodelstransfer,li2023vision,li2024cogact,liu2025hybridvlacollaborativediffusionautoregression,zheng2025tracevlavisualtraceprompting,zhen20243dvla3dvisionlanguageactiongenerative,zhang2025groundingactionscameraspace} own the capability to make action policy according to task instruction through continuous observations or other interactions with the environment. This robotic manipulation capability is developed from training strategies such as imitation learning (IL) through annotated expert demonstrations performing the same tasks. Also equipped with Multimodal Large Language Model (MLLM) with the basic spatial understanding of the world through pre-training, it is difficult for these VLA models to always present feasible manipulation trajectories solely through the observed 2D images. And the multiview generalization ability of these VLA models also degrades when these observed 2D images differ in perspectives, heights, backgrounds, or other aspects from those in the learning stage of these models. Although works like \cite{brohan2023rt2visionlanguageactionmodelstransfer} and \cite{wu2023unleashinglargescalevideogenerative,cheang2024gr2generativevideolanguageactionmodel} release dependency on in-domain (ID) data request during fine-tuning by pretraining the MLLM backbones of their VLA models with huge amount of internet-scale data via co-finetuning and transfer learning, experiments in most of the aforementioned studies indicate that data homogeneity remains the indispensable prerequisite for robust task execution. In plain words, if the visual characteristics of the environment observed at the deployment stage are not the same as the training datasets, the VLA models are more likely to struggle to accomplish the tasks.

In light of this, we propose an approach to decouple this similarity requirement by enhancing the generalization ability across multiple perspectives relying only on 2D observed images, as Figure \ref{lv-i0} shows. Specifically, we focus on filling the perspective gap by properly fusing diverse views. In contrast to solutions that augment training data with extra collected views \cite{wen2025diffusionvlageneralizableinterpretablerobot} or those that reconstruct 3D scenes from sparse viewpoints \cite{brohan2023rt1roboticstransformerrealworld,brohan2023rt2visionlanguageactionmodelstransfer}, we draw inspiration from \cite{liu2023llava,liu2023improvedllava} to design a lightweight MLP-based fusion module for multiview feature alignment. This module injects aligned information from out-of-domain (OOD) perspective data into the VLA model in its latent space through fine-tuning, thus assigning the VLA model the trajectory generation capability not only from the single viewpoint but also across previously unseen ones.
By means of latent feature visualization and experiments in both simulated and real-world setup, we demonstrate the advantage of our proposed lightweight solution in releasing the constraint of perspective consistency between ID and OOD data for VLA models.
In summary, our contributions are primarily as follows:
\begin{itemize}

    \item We for the first time implement a lightweight framework VLA-LPAF which relies on latent perspective feature fusion using only 2D images and make the VLA models less constrained to the perspective homogeneity between training stages and deployment.

    \item We instantiate VLA-LPAF with RoboFlamingo \cite{li2023vision} to construct RoboFlamingo-LPAF, and show its effectiveness of it through extensive experiments on multiple simulated datasets and real-world tasks.

\end{itemize}

\label{sec:intro}


\section{Related Work}

\begin{figure}[!ht]
\centering
\captionsetup{justification=justified, singlelinecheck=false}
\includegraphics[width=0.95\columnwidth]{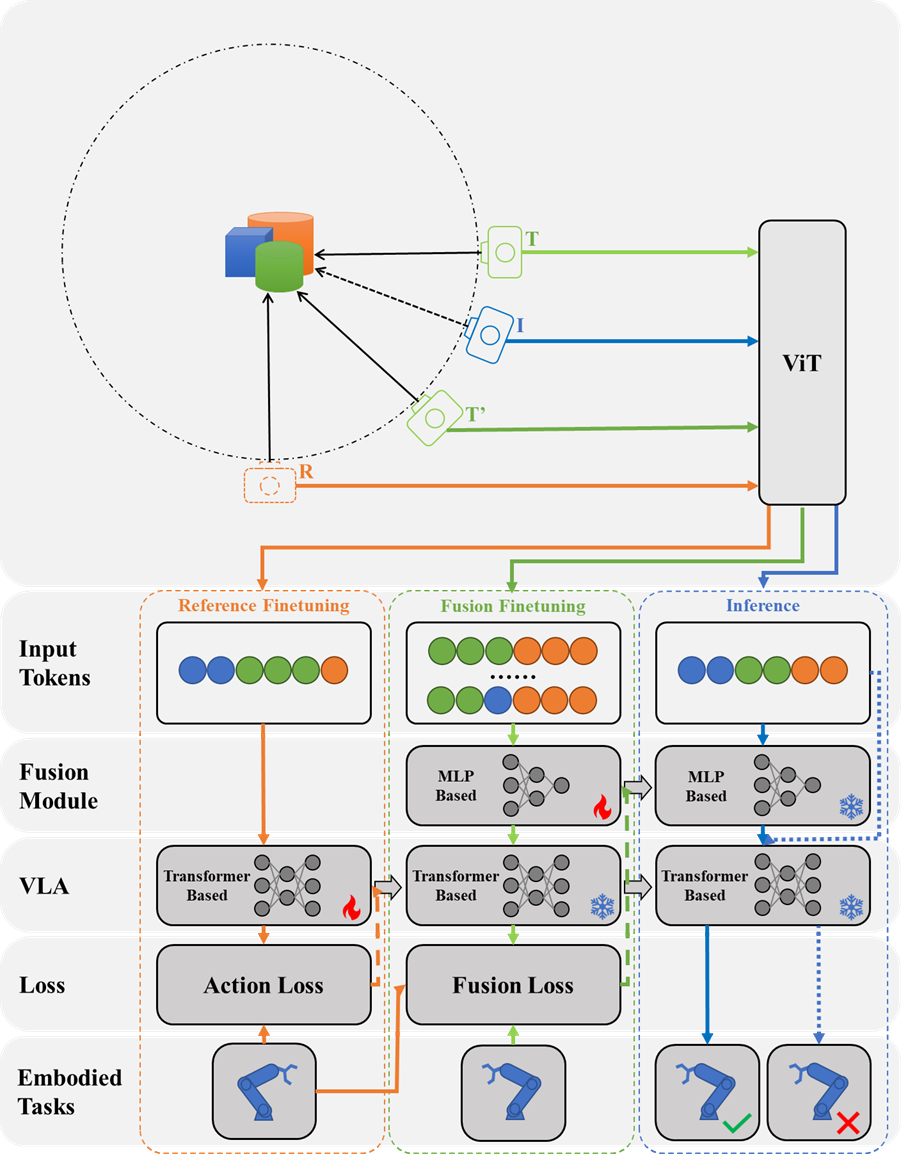}
\caption{The view generalization capability of the proposed VLA-LPAF is displayed through the reference finetuning and fusion finetuning sections illustrated in this figure. In contrast to common VLA models, our approach incorporates an MLP-based fusion module that efficiently injects aligned multi-view information in the latent space. The comparison between the solid blue line and the dashed line during inference highlights the advantage.}
\label{lv-i1}
\end{figure}

The VLA has increasingly emerged as a prevailing framework for action trajectory generation in the development of recent robotic generalist. 

Based on the action generation mechanism, VLA models can be categorized into three types: the autoregressive, the diffusion (or flow matching) and the hybrid. The autoregressive VLA models like RT-2 \cite{brohan2023rt2visionlanguageactionmodelstransfer}, RoboFlamingo\cite{li2023vision} and OpenVLA\cite{kim2024openvlaopensourcevisionlanguageactionmodel} encode visual and text inputs before feeding them to the Large Language Models (LLM) for causal actions generation in an autoregressive manner similar to the way the LLMs generate text. The diffusion-based approaches such as Diffusion Policy \cite{chi2024diffusionpolicy}, 3D Diffusion Policy \cite{Ze2024DP3}, and 3D Diffuser Actor \cite{ke20243ddiffuseractorpolicy} condition the diffused action generation process on latent features extracted from encoded and aligned multimodal inputs. The hybrid VLA models decouple the comprehension capability of MLLMs from the action generation and finetune them respectively. Typical works in this category include CogAct \cite{li2024cogact}, Diffusion-VLA \cite{wen2025diffusionvlageneralizableinterpretablerobot}, Hybrid-VLA \cite{liu2025hybridvlacollaborativediffusionautoregression}, and $\pi$0 \cite{black2024pi0visionlanguageactionflowmodel} and $\pi$0.5 \cite{intelligence2025pi05visionlanguageactionmodelopenworld}.

In terms of the modalities of visual inputs, VLA models can be classified into two categories: the 2D modality and the 3D modality (the 2.5D data such as depth maps are viewed as the 3D modality for convenience of discussion). The former category takes only 2D images from a static third-person viewpoint camera to present the global perspective of the environment and several in-wrist cameras to provide local details. Some VLA models \cite{brohan2023rt1roboticstransformerrealworld,brohan2023rt2visionlanguageactionmodelstransfer} require temporally multiple images, while others like \cite{zheng2025tracevlavisualtraceprompting} render 2D images using additional visualized information. For VLA models that need 3D modality input \cite{li2025bridgevlainputoutputalignmentefficient,zhen20243dvla3dvisionlanguageactiongenerative}, the pipeline is commonly introduced to the unified 3D coordinate system and processed with the corresponding 3D encoding, position embedding and feature extraction methods.

Multiview images are defined as the combination of both global and local observations of the environment acquired by cameras at different positions in most of the VLA solutions. Taking the single-arm embodied device as an example: the camera responsible for the global observation is usually fixed at positions capable of offering top or side view of the entire environment. The camera or cameras for the local observations are generally mounted on the movable end-effector of the embodied device. These images are encoded and concatenated before being processed later. The researches to address the issue of generalization performance degradation due to perspective misalignment are listed in the following part:

\begin{itemize}

    \item \textbf{Through Data Collection:}  Solutions such as \cite{yan2024robommallinonemultimodallarge} perform datasets collection and merge based on 3D unification, which requires extensive efforts to convert the dimensionality of these datasets. Among the few works that try to solve this problem in the aspect of 2D modality, \cite{wen2025diffusionvlageneralizableinterpretablerobot} just collects observed images through additional acquisition in more viewpoints to improve the model's view-adaptivity. 
    
    \item \textbf{Through Data Re-rendering:} VLA architectures that take multi-view images as input like \cite{goyal2023rvtroboticviewtransformer,goyal2024rvt2learningprecisemanipulation,li2025bridgevlainputoutputalignmentefficient} obtain additional perspectives of the observation from virtual viewpoints through re-rendering of the reconstructed 3D pointcloud of the environment. These approaches usually employ off-the-shelf RGBD cameras to get depth information and perform 3D reconstruction of the scenes. Virtual cameras are then placed within according to the specific number and view-angle requirements to offer the supplemented perspectives.
    
    \item \textbf{Our Solution:} Our proposed VLA-LPAF framework generally improves the generalization capability of VLA models through efficient fusion of perspective information as Figure \ref{lv-i1} shows. Unlike \cite{wen2025diffusionvlageneralizableinterpretablerobot}, which proportionally reduces the number of images captured at the original viewpoint when images at additional viewpoints are added, we retain the perspective diversity through feature alignment in the latent space and our ablation experiments prove its effectiveness. In terms of information fusion, \cite{zhang2025groundingactionscameraspace} employs a similar hardware setup of the system that integrates the coordinate systems of robotic arms from multiple views into a unified camera-centric coordinate system. This approach releases the computation burden on VLA models in mapping actions through 2D images. However, it differs from our pure 2D-based fusion method conducted in the latent space. VLA schemes based on 3D modal inputs like \cite{goyal2023rvtroboticviewtransformer,goyal2024rvt2learningprecisemanipulation,qu2025spatialvlaexploringspatialrepresentations,sun2025geovlaempowering3drepresentations} also apply fused multiple perspective images to enhance spatial reasoning ability by means of re-rendering techniques. Compared to them, our proposed VLA-LPAF presents a more lightweight framework involving only 2D data.

\end{itemize}

\label{sec:related}

\section{Method}

\subsection{Problem Definition}

\begin{figure*}[!ht]
\centering
\captionsetup{justification=justified, singlelinecheck=false}
\includegraphics[width=1.9\columnwidth]{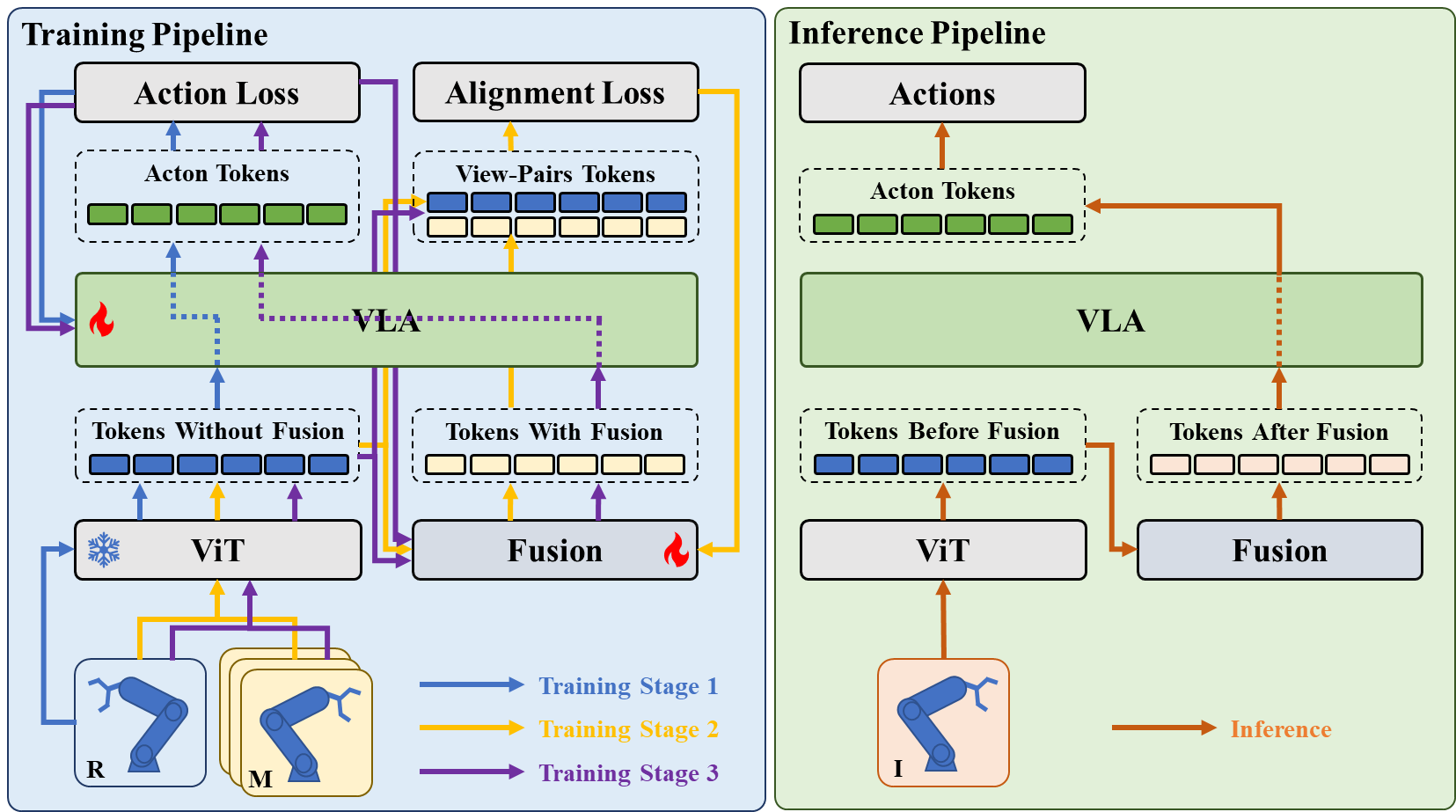}
\caption{We design a three-staged training procedure that are represented by arrows of different colors in the left half of this figure. In the first stage, the reference perspective images (denoted as R images from reference perspective) are utilized for supervised fine-tuning (SFT) of the VLA model exclusively through action loss. The second stage aligns and fuses the R images with the auxiliary perspectives images (denoted as M from multiple auxiliary perspectives) in the encoded latent space, performing SFT on the fusion module via alignment loss. The third stage conducts simultaneous SFT on both the fusion module and the VLA model. The inference procedure are depicted in the right half as the fine-tuned VLA-LPAF model is able to handle observed images (denoted as I images) regardless of perspective consistency constrain.}
\label{lv-i2}
\end{figure*}

The VLA model can be generally viewed as a policy model $\pi_\theta(a_t \mid o_t, l)$ with a backbone LLM $\pi_\theta$ that maps the multi-modal combination of language instruction $l$ and images $o_t$ observed at time $t$ to the future executable action $a_t$ or action chunk $\{a_t\}_{t=0}^T$. For brevity, we assume that the policy model $\pi_\theta$ generates a single action $a_t$ at a time. The $i$-th trajectory for the $s$-th task is thus comprised of the actions, the text instruction and the observed images at each timestep, which is denoted as $\tau_i = \{a_{t_s}, o_{t_s}, l_s\}_{t_s=0}^{T_s}$. We treat all the $S$ example tasks' trajectories in the training dataset $\tau=\{\{a_{t_s}, o_{t_s}, l_i\}_{t_s=0}^{T_i}\}_{i=0,s=0}^{I,S}\in \mathbb{D}_{R}$ as expert behaviors that the VLA model should clone and iteratively optimize the network parameters $\theta$ in $\pi_\theta(a_t \mid o_t, l)$ according to the following loss function:
\begin{equation}
\label{lv-eq0}
\begin{split}
\mathbb{L}_{action}&=\mathbb{E}_{\tau \sim \mathbb{D}_{R}}[\sum\limits_{s=0}^{S}\sum\limits_{t_s=0}^{T_{t_s}} log(\pi_\theta( \tau_i))]\\
&=\mathbb{E}_{(a, o, l) \sim D}[\sum\limits_{s=0}^{S}\sum\limits_{t_s=0}^{T_{t_s}} log(\pi_\theta(a_t|o_t, l))]
\end{split}
\end{equation}

If the global camera is installed in different positions between the training dataset $\mathbb{D}_{R}$ used for fine-tuning and the test set $\mathbb{D}_{test}$ for deployment, the discrepancy in visual features between the images $\{o_t^M\}_{t=0}^{T_i}$ observed from other viewpoints in the dataset and the images $\{o_t^R\}_{t=0}^{T_i}$ in the training set will puzzle $\pi_\theta(a_t \mid o_t, l)$ and consequently results in a decrease in precision for these OOD cases. \cite{wen2025diffusionvlageneralizableinterpretablerobot} addresses this issue by adding $\mathbb{D}_{R}$ (referred to as reference perspective later for brevity) with 2D image data captured from those OOD perspective similar to those in $\mathbb{D}_{test}$ (referred to as auxiliary perspective later for brevity) and fine-tuning $\pi_\theta(a_t \mid o_t, l)$ utilizing the mixed multi-view dataset $\mathbb{D}_M$ according to Equation \ref{lv-eq1} to mitigate this problem:
\begin{equation}
\label{lv-eq1}
\begin{split}
\mathbb{L}_{action}=\mathbb{E}_{(a, o, l) \sim \mathbb{D}_{M}}[\sum\limits_{s=0}^{S}\sum\limits_{t_s=0}^{T_{t_s}} log(\pi_\theta(a_t|o_t^M, l))]
\end{split}
\end{equation}

To ensure training efficiency, \cite{wen2025diffusionvlageneralizableinterpretablerobot} proportionally reduces the number of visual feature tokens to balance the total size of the dataset. As mentioned earlier, this manipulation improves the perspective adaptivity of OOD cases at the cost of sacrificing the average generalization performance, which we will prove through ablation experiments. In contrast, our VLA-LPAF constructs a lightweight learnable MLP-based fusion module $\mathbb{F}_{\theta'}$ that aligns the latent features extracted from multiview observation images $\{o_t^M\}$ as Equation \ref{lv-eq2} describes:
\begin{equation}
\label{lv-eq2}
\begin{split}
\hat{\mathbb{L}}_{action}=\mathbb{E}_{\tau \sim \mathbb{D}_M}[\sum\limits_{t=0}^{T_i} log(\pi_\theta(a_t|\mathbb{F}_{\theta'}(o_t^M), l))]
\end{split}
\end{equation}
We will also validate the advantage of the proposed VLA-LPAF in the ablation experiment section.

\subsection{Architecture Overview}
In addition of the indispensable components of common VLA models, our VLA-LPAF framework is composed of an additional 2D perspective fusion module. During the training phase of VLA-LPAF, we sequentially employ a single-view dataset $\mathbb{D}_R$ containing only the reference perspective together with a multi-view dataset $\mathbb{D}_M$ containing other auxiliary perspectives. According to Equation \ref{lv-eq2}, we fine-tune the backbone LLM $\pi_\theta$ and the fusion module $\mathbb{F}_{\theta'}$ respectively in different stages. In the inference stage, the observed 2D images (containing both a global and an in-wrist images) set $\{o_t^{G,L}\}$ is encoded, concatenated and fused into a latent space aligned to the reference perspective before being sent into downstream modules for action generation. The general training and inference pipeline of VLA-LPAF is illustrated in Figure \ref{lv-i2}.

\subsection{Alignment through Perspective Fusion in Latent Space}
The core of VLA-LPAF is the fusion module that performs alignment according to the latent visual features between 2D images of the reference perspective and the auxiliary perspectives. To maintain the lightweight characteristics of the entire framework, we apply computationally efficient MLP structures to build the fusion module. As the lack of depth modality, we select the latent features extracted by the ViT encoder and make the MLP-based to implicitly develop a proper correlation between reference and auxiliary features.
\begin{figure}[!ht]
\centering
\captionsetup{justification=justified, singlelinecheck=false}
\includegraphics[width=0.95\columnwidth]{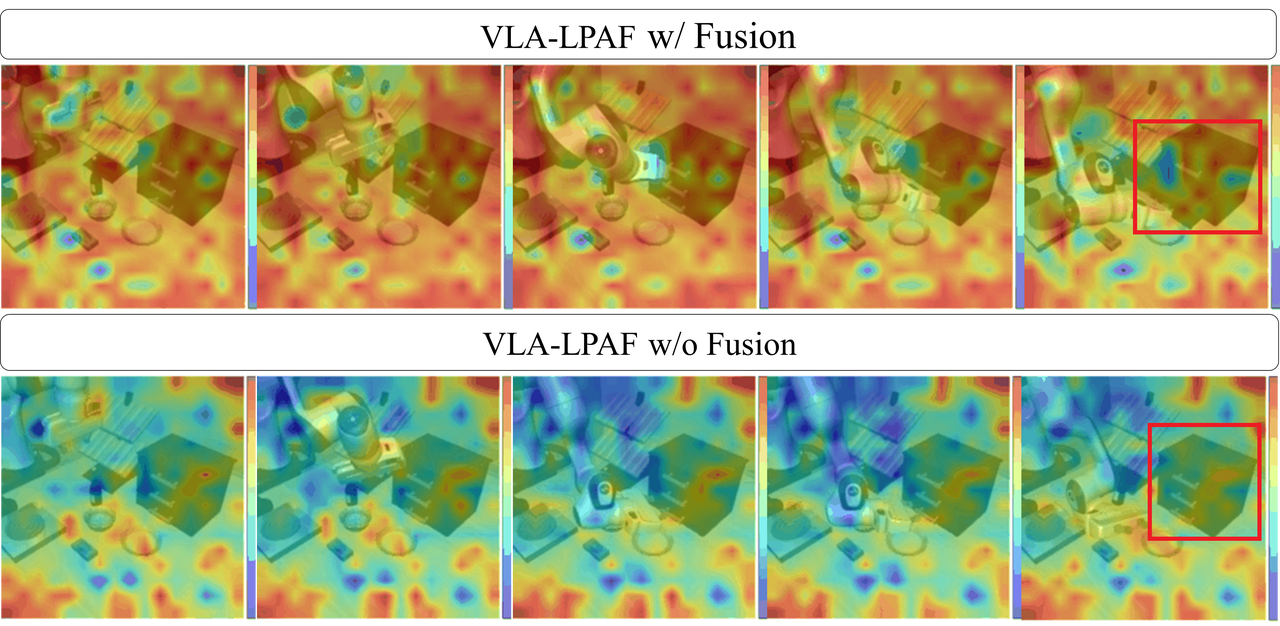}
\caption{We visualize in heatmap the latent feature similarity of the tokenized observation with and without our fusion module. The red region means high similarity while the blue means the opposite. The fact that the main target object (the drawer as we mark using a red rectangle) retains in red processed by our fusion module demonstrates its effectiveness.}
\label{lv-i3}
\end{figure}

We also prove the effectiveness of our fusion module through visualization. As illustrated in the comparative heatmaps shown in Figure \ref{lv-i3}, we can observe that the regions represented by the activated parameters in the fusion module are correctly related to objects that are also closely relevant to the execution of embodied tasks from different viewpoints.

\subsection{Training Strategy}
We devise a three-stage training pipeline for the proposed VLA-LPAF to ensure that each major component of the framework is trained adequately and thoroughly:
\begin{itemize}
    \item \textbf{The Single-View Stage for Action Only:} In this stage, we follow the standard VLA fine-tuning principle. For all tasks, we only involve the observations from $\mathbb{D}_R$ with the task instruction $l$ as inputs, and train the policy model $\pi_\theta$ supervised by Equation \ref{lv-eq0}. We freeze the ViT backbone during this stage and fine-tune only the LLM parameters $\theta$.
    
    \item \textbf{The Multi-View Stage for Fusion Only:} We train the fusion module in this stage to foster the viewpoint alignment capability using both $\mathbb{D}_R$ and $\mathbb{D}_M$. To preserve the lightweight characteristic of VLA-LPAF, we introduce no extra tools or modalities except for the original 2D images. Consequently, we cannot accurately align image spaces across heterogeneous viewpoints by camera parameters according to the pinhole model such as \cite{qu2025spatialvlaexploringspatialrepresentations} or \cite{yan2024robommallinonemultimodallarge}. Inspired by the conclusion of \cite{liu2023llava} that a network composed of several MLP layers can efficiently project visual features to the linguistic feature space in the latent space, we adopt a similar architecture that performs the alignment between these encoded observed images from different perspectives. During this stage, we only train the parameters of the fusion module $\theta'$ with the loss described in Equation \ref{lv-eq4}.
    \begin{equation}
    \label{lv-eq4}
    \begin{split}
    \mathbb{L}_{alignment}=\frac{1}{N}\sum\limits_{n=0}^{N} |ViT(o^R_n)-\mathbb{F}_{\theta'}(ViT(o^M_n))|_2
    \end{split}
    \end{equation}
    In detail, we apply a progressive strategy to accomplish the training for the fusion module that incrementally incorporates multiview data from $\mathbb{D}_M$ to reduce the learning burden on the alignment module. Our subsequent ablation studies further demonstrate that this approach yields better performance than exposing all data to the fusion module.
    
    \item \textbf{The Multi-View Stage for Both Action and Fusion:} In this stage, we integrate the capabilities learned in the two aforementioned stages and simultaneously adjust $\theta$ and $\theta'$ through both action loss and alignment loss as Equation \ref{lv-eq5} shows:
    \begin{equation}
    \label{lv-eq5}
    \begin{split}
    \mathbb{L}_{all}=\mathbb{\hat{L}}_{action}+\mathbb{L}_{alignment}
    \end{split}
    \end{equation}
    
\end{itemize}

\section{Experiments}

The primary objective of this paper is to evaluate whether our VLA-LPAF framework plays a vital role in the development of the multiview adaptivity from 2D images alone. We prove this by instantiation of the proposed framework with a specific VLA model and seek to answer these following questions through our experiments:

(1) Does VLA-LPAF help VLA break from the constraints of viewpoint consistency?

(2) To what extent are the key modules in VLA-LPAF optimized?

(3) Does the unconstrained generalization ability provided by VLA-LPAF still hold across different datasets?

\subsection{Implementation Details}
To answer the questions below, we first instantiate the VLA-LPAF framework with RoboFlamingo \cite{li2024visionlanguagefoundationmodelseffective}, generating the VLA model RoboFlamingo-LPAF to perform our experiments. We fine-tune it on a cluster of 8 NVIDIA A800 80 GB GPUs using CALVIN \cite{liu2023libero}, LIBERO \cite{mees2022calvin} and our customized CabinEnv simulated datasets. Based on a simulated cabin environment, CabinEnv includes two tasks: pressing buttons and flipping the levers.
\begin{figure}[!ht]
\centering
\captionsetup{justification=justified, singlelinecheck=false}
\includegraphics[width=0.95\columnwidth]{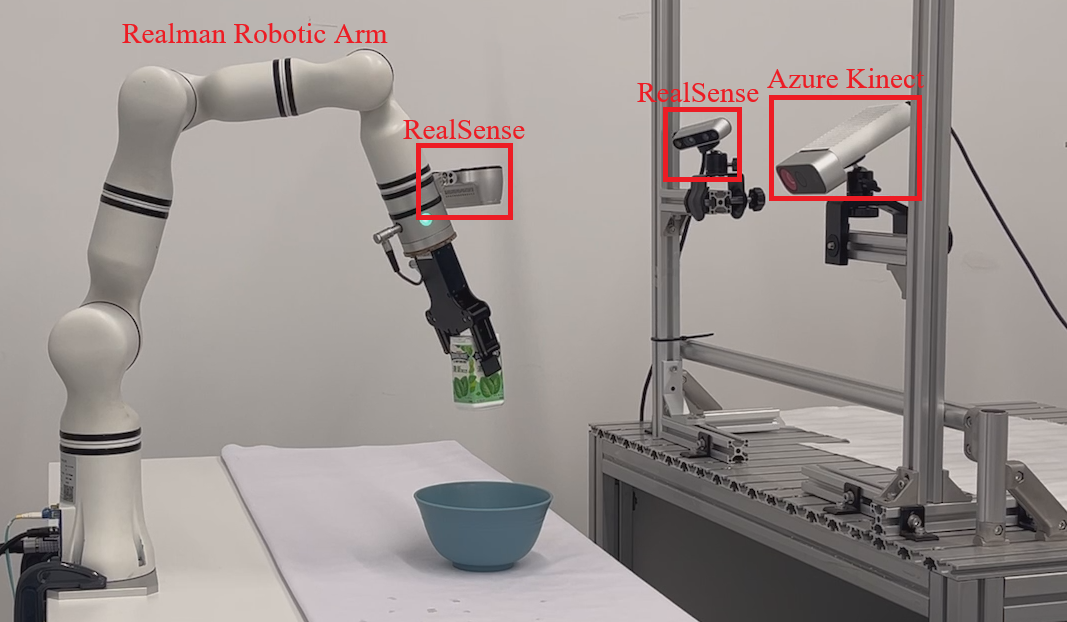}
\caption{Our real world tasks validation based on the Realman robotic single-arm and several RGB-D cameras.}
\label{lv-i4}
\end{figure}

As shown in Figure \ref{lv-i4}, we set up a single-arm robotic platform using the Realman RML series robotic arm and deploy the RoboFlamingo-LPAF model to evaluate its performance in real world tasks. We apply the Intel RealSense D415 as the global reference and the in-wrist 2D observing cameras. The Azure Kinect are used as the global auxiliary 2D observing camera. The input images from different viewpoints are resized to a unified resolution of $224\times224$. Since we focus solely on validating the model’s multiview adaptability.

\subsection{Main Results}

\begin{itemize}

    \item \textbf{Validation in Simulated Environment:} We fine-tune and evaluate RoboFlamingo-LPAF and the original RoboFlamingo as baseline model with similar amount of multiview perspective data in $\mathbb{D}_R$ and $\mathbb{D}_M$ for a fair comparison. We fine-tune the baseline following the same data addition policy from \cite{wen2025diffusionvlageneralizableinterpretablerobot}.
    
    Specifically, we select the reference perspective and regard it as 0°. From the reference viewpoint, $j$ trajectories are sampled for each task to form $\mathbb{D}_R$. During the data collection phase for fine-tuning, we set the other viewpoints clockwise and counterclockwise multiple times at a fixed angular interval around the reference perspective to generate a total of $v$ auxiliary perspectives. For each auxiliary perspective, $s$ tasks are performed with $j$ trajectories collected for each task to construct $\mathbb{D}_M$. The fine-tuning datasets $\mathbb{D}_R$ and $\mathbb{D}_M$ equally contain $(v+1) \times s \times j$ trajectories for model fine-tuning. For the sake of fair comparison, we also sample $(v+1) \times s \times j$ trajectories solely from the reference viewpoint to train the baseline model. For the CALVIN dataset, the parameters are set to: $a=45°$, $v=4$, $s=5$, and $j=28$. For the LIBERO-Goal dataset, the parameters are set as: $a=45°$, $v=4$, $s=6$, and $j=5$. For the CabinEnv dataset, we set the parameters as: $a=45°$, $v=4$, $s=2$, and $j=200$. During the data collection phase for the test, the viewpoints are set clockwise and counterclockwise 9 times each with an interval of $10°$. The comparative results on the three different simulated datasets are presented in the following figures respectively.

    It can be concluded that the proposed RoboFlamingo-LPAF model achieves better performance than baseline model on CALVIN, LIBERO-Goal and our customized CabinEnv. 
    
    \begin{figure}[!ht]
    \centering
    \captionsetup{justification=justified, singlelinecheck=false}
    \includegraphics[width=0.95\columnwidth]{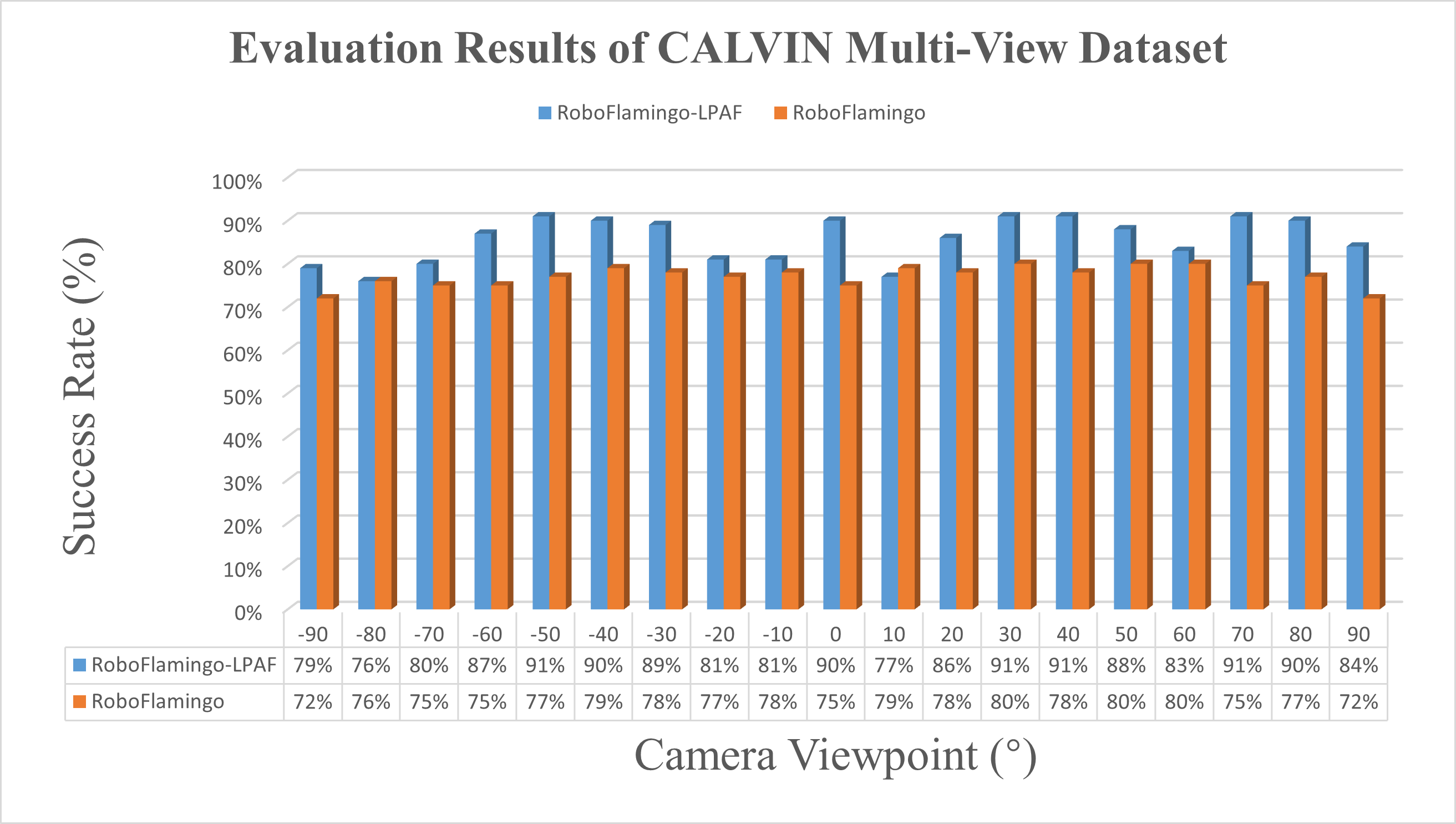}
    \caption{Performance comparison results on multiview CALVIN dataset.}
    \label{lv-i5}
    \end{figure}
    
    \begin{figure}[!ht]
    \centering
    \captionsetup{justification=justified, singlelinecheck=false}
    \includegraphics[width=0.95\columnwidth]{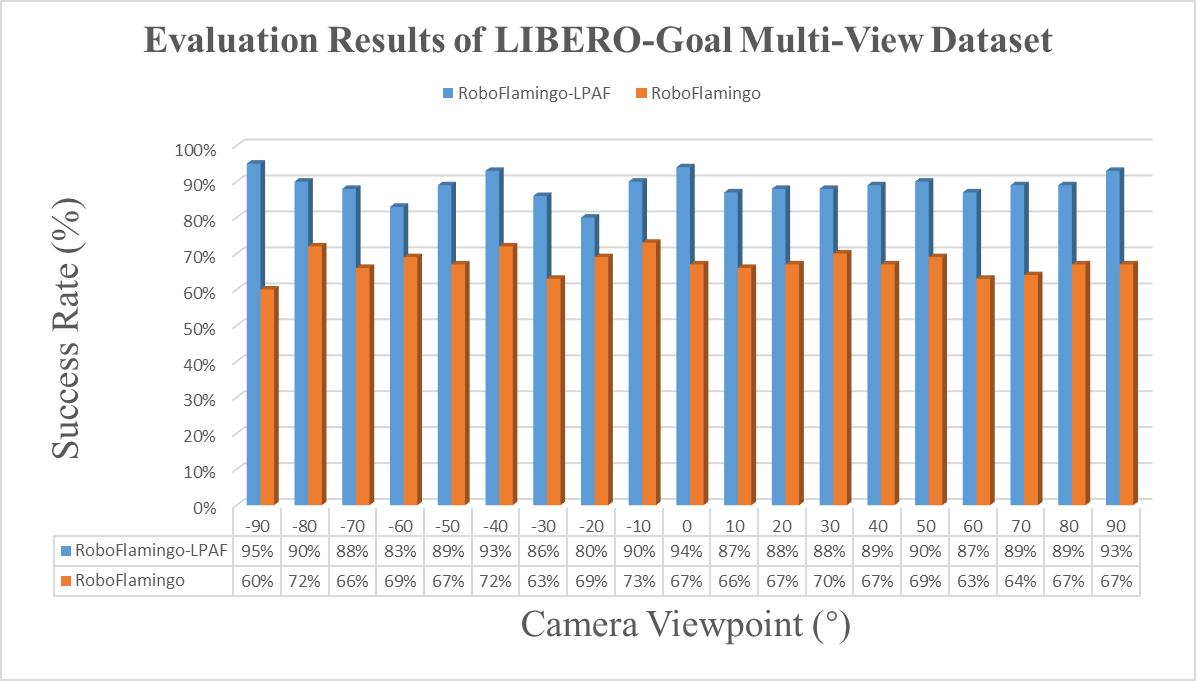}
    \caption{Performance comparison results on multiview LIBERO-Goal dataset.}
    \label{lv-i6}
    \end{figure}
    
    \begin{figure}[!ht]
    \centering
    \captionsetup{justification=justified, singlelinecheck=false}
    \includegraphics[width=0.95\columnwidth]{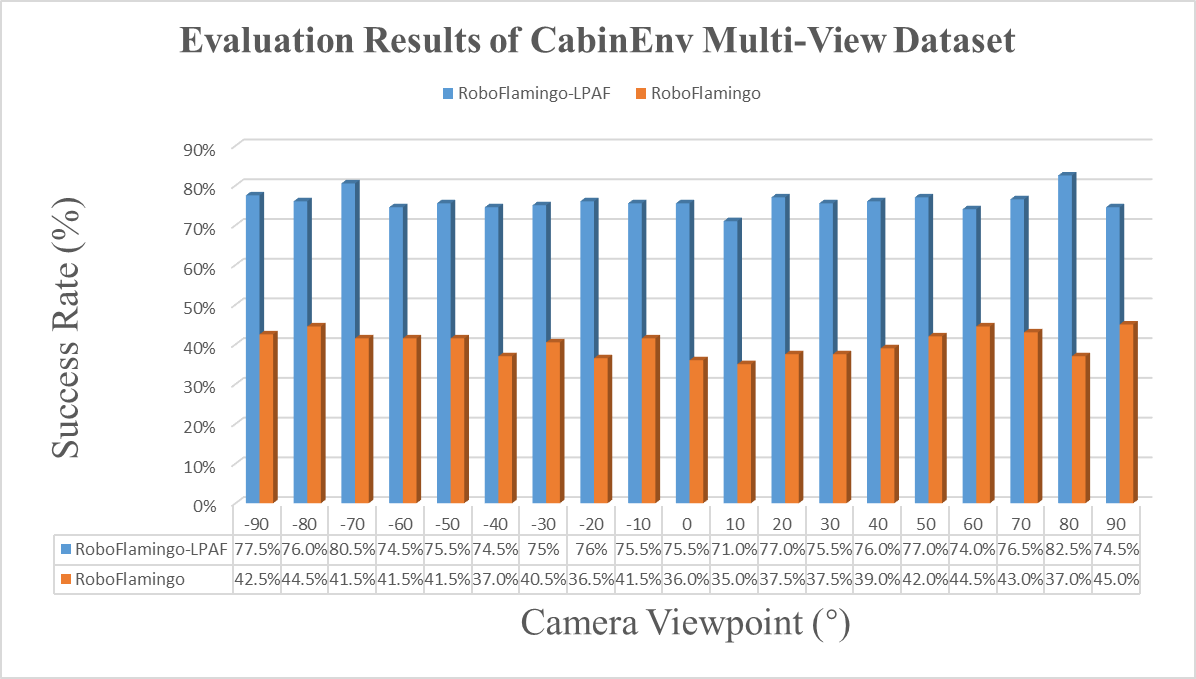}
    \caption{Performance comparison results on multiview customized CabinEnv dataset.}
    \label{lv-i7}
    \end{figure}
    
    \item \textbf{Validation in Real World:} We also compare RoboFlamingo-LPAF to baseline model in terms of four real-world tasks. Figure \ref{lv-i5} illustrates the comparison:
    
    \begin{figure*}[!ht]
    \centering
    \captionsetup{justification=justified, singlelinecheck=false}
    \includegraphics[width=1.9\columnwidth]{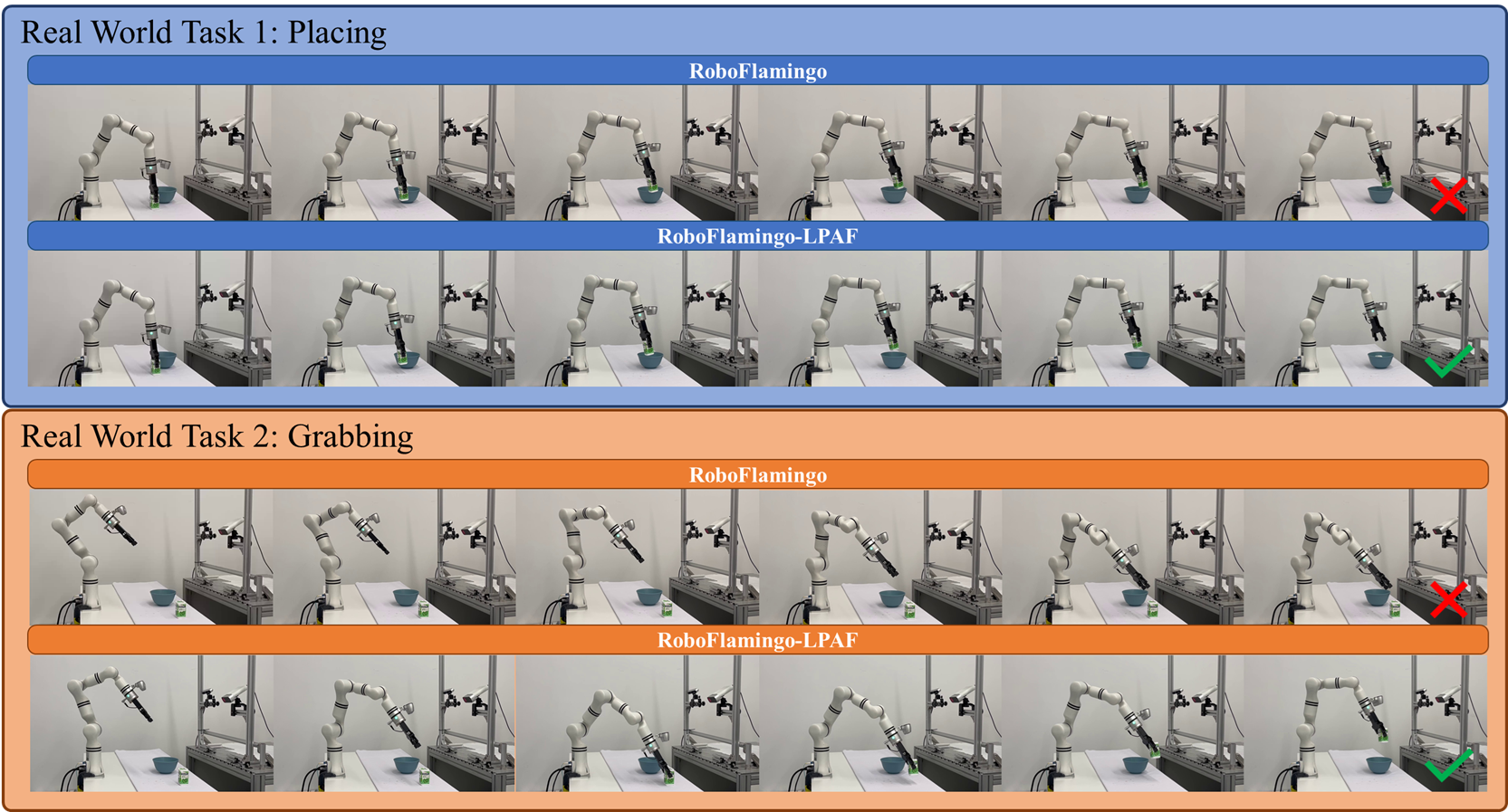}
    \caption{Comparison of multiview tasks execution using RoboFlamingo-LPAF and baseline models. We choose the $30°$ viewpoint to perform the inference, which is not included in $\mathbb{D}_M$. RoboFlamingo-LPAF successfully completes these tasks while the original RoboFlamingo fails..}
    \label{lv-i8}
    \end{figure*} 
    
    According to the validation results from the simulation environment and the real-world tasks, we can present affirmative answer to the questions (1) and (3) with the advantageous perspective generalization capability of our RoboFlamingo-LPAF.

\end{itemize}

\section{Ablation Experiments}
To answer question (2), we conduct ablation studies on the training pipeline, loss configuration and major components of our RoboFlamingo-LPAF to prove its well-roundness. All subsequent experiments are performed on tasks from the LIBERO-Goal dataset.

\subsection{Alignment Loss Item}
\begin{figure}[!ht]
\centering
\captionsetup{justification=justified, singlelinecheck=false}
\includegraphics[width=0.95\columnwidth]{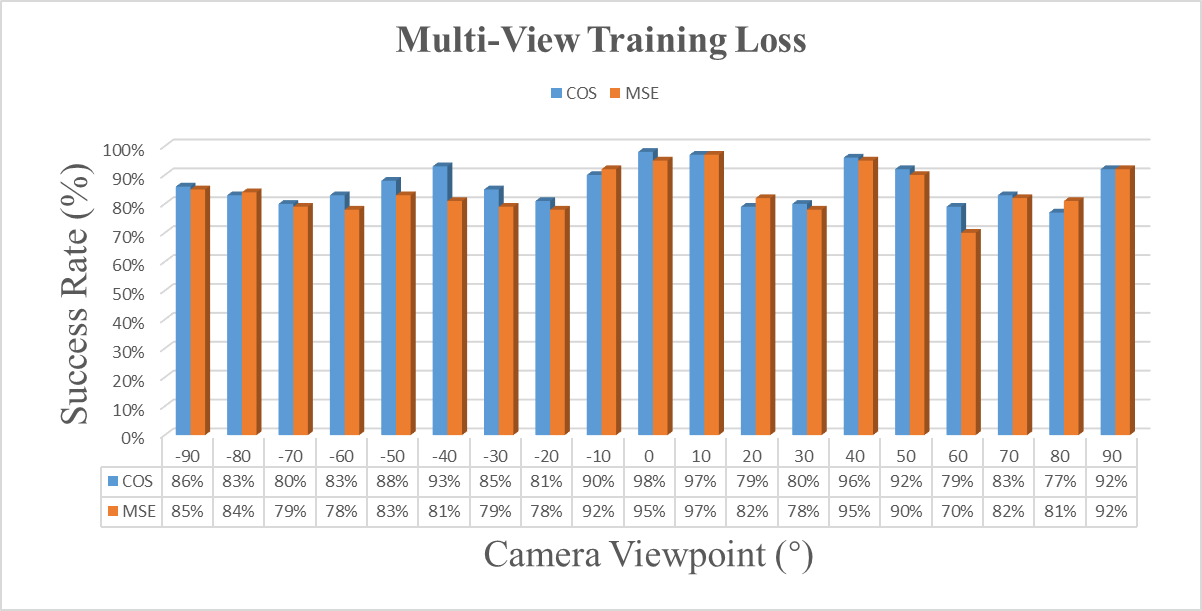}
\caption{Ablation experiment comparison results on different alignment loss items.}
\label{lv-i9}
\end{figure}

For the latent perspective feature alignment loss to supervise the neural parameters $\theta'$ in the fusion module, we compare the Mean Squared Error (MSE) based method to the Cosine Similarity (COS) based method. Table \ref{lv-t0} shows the corresponding average task success rates for both cases:
\begin{table}[!ht]
    \centering
    \scalebox{1.0} {
    \renewcommand{\arraystretch}{1.29}
        \begin{tabular} {c|c}
        \toprule
        Alignment Loss Item & Mean Success Rate \\
        \midrule
        \textbf{COS} & \textbf{86.42\%} \\
        MSE & 84.26\% \\
        \bottomrule
        \end{tabular}
    }
    \captionsetup{justification=justified, singlelinecheck=false}
    \caption{Ablation experiment on different alignment loss items in RoboFlamingo-LPAF.} 
    \label{lv-t0}
\end{table}
Figure \ref{lv-i9} demonstrates this from the results from other validation viewpoints.
    
\subsection{Multiview Data Involvement Strategy}
In case the number of auxiliary viewpoints is more than two during the fusion training stage, we notice that different strategies for introducing auxiliary viewpoints plays a significant role in the final task success rate. We evaluate two strategies: a one-pass addition approach that incorporates the data of all the auxiliary views simultaneously and a progressive approach that involves these auxiliary data gradually. As indicated in Table \ref{lv-t1}, the progressive strategy enables the fusion network to integrate and align multiview information more effectively.
\begin{table}[!ht]
    \centering
    \scalebox{1.0} {
    \renewcommand{\arraystretch}{1.29}
        \begin{tabular} {c|c}
        \toprule
        Involvement Strategy & Mean Success Rate \\
        \midrule
        w/o gradually & 86.42\% \\
        \textbf{w/ gradually} & \textbf{87.79\%} \\
        \bottomrule
        \end{tabular}
    }
    \captionsetup{justification=justified, singlelinecheck=false}
    \caption{Ablation experiment on different auxiliary perspectives data addition strategies in RoboFlamingo-LPAF.} 
    \label{lv-t1}
\end{table}
Figure \ref{lv-i10} gives a more well-round comparison viewpoint-wisely:
\begin{figure}[!ht]
\centering
\captionsetup{justification=justified, singlelinecheck=false}
\includegraphics[width=0.95\columnwidth]{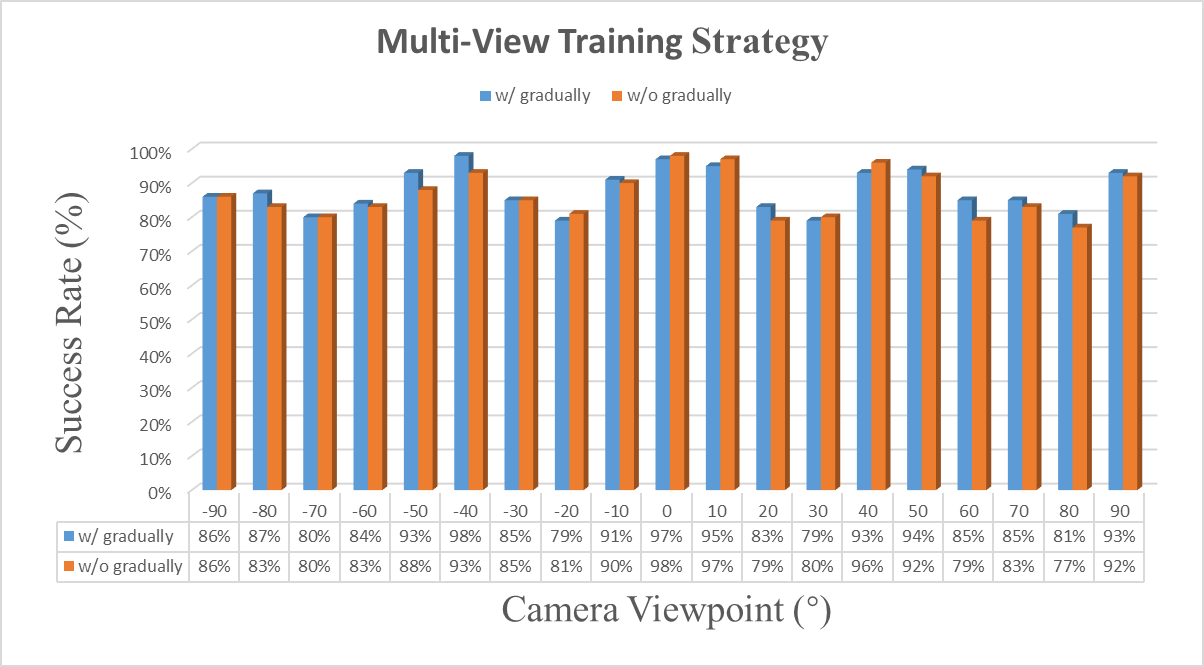}
\caption{Ablation experiment comparison results on different multiview data involvement strategies.}
\label{lv-i10}
\end{figure}

\subsection{Parameters Updating Strategy}
In the multiview stage (the third stage in our three-stage training recipe) for both action and fusion, we compare the finetuning strategy of updating the neural parameters in the backbone LLM $\theta$ with $\theta'$ in the fusion module with the strategy that the $\theta$ are frozen. As shown in Table \ref{lv-t2}, the former strategy during this stage leads to better average performance and in most of the test viewpoints.
\begin{table}[!ht]
    \centering
    \scalebox{1.0} {
    \renewcommand{\arraystretch}{1.29}
        \begin{tabular} {c|c}
        \toprule
        Involvement Strategy & Mean Success Rate \\
        \midrule
        w/ freeze & 88.63\% \\
        \textbf{w/o freeze} & \textbf{88.84\%} \\
        \bottomrule
        \end{tabular}
    }
    \captionsetup{justification=justified, singlelinecheck=false}
    \caption{Ablation experiment on different parameters updating strategies in RoboFlamingo-LPAF.} 
    \label{lv-t2}
\end{table}

The experimental results of these ablation studies above illustrate that we adopt the strategies that most benefit task completion for all key modules in the pipeline of our RoboFlamingo-LPAF, which can be viewed as a confident response to the question (2).

\label{sec:method}

\section{Limitation}
Although RoboFlamingo-LPAF is able to present advantageous performance across a variety of simulated benchmarks and real-world tasks, the following limitations exist in our proposed VLA-LPAF framework: (1) VLA-LPAF needs to be instantiated with more VLA models and tested by more benchmarks; (2) The fusion module can be implemented in a more effective form without compromising efficiency. These provide promising directions for our future research.

\section{Conclusion}
This paper investigates the fusion of latent visual representations from various perspectives to enhance the multiview generalization capability of VLA models via a lightweight framework VLA-LPAF. Through the instantiation of VLA-LPAF with the common VLA model like RoboFlamingo without any extra higher dimensional data, we achieve better task success rate in both simulated benchmarks and real-world platform.
\label{sec:conclusion}

{\small
\bibliographystyle{ieeenat_fullname}
\bibliography{11_references}
}


\end{document}